\documentclass{article}

\usepackage[final, nonatbib]{nips}

\usepackage[utf8]{inputenc} % allow utf-8 input
\usepackage[T1]{fontenc}    % use 8-bit T1 fonts
\usepackage{hyperref}       % hyperlinks
\usepackage{url}            % simple URL typesetting
\usepackage{booktabs}       % professional-quality tables
\usepackage{amsfonts}       % blackboard math symbols
\usepackage{nicefrac}       % compact symbols for 1/2, etc.
\usepackage{microtype}      % microtypography
\usepackage{multirow}
\usepackage{array}
\usepackage{graphicx}
\usepackage{color}
\usepackage{fancyhdr}
\title{CMSIS-NN: Efficient Neural Network Kernels for Arm Cortex-M CPUs}

\author{
  Liangzhen Lai \\
  Arm Inc. \\
  \texttt{liangzhen.lai@arm.com} \\
  \And
  Naveen Suda \\
  Arm Inc.\\
  \texttt{naveen.suda@arm.com} \\
  \And
  Vikas Chandra\\
  Arm Inc.\\
  \texttt{vikas.chandra@arm.com} \\
}

\begin{document}

\maketitle

\begin{abstract}
Deep Neural Networks are becoming increasingly popular in 
always-on IoT edge devices performing data analytics right at the
source, reducing latency as well as energy consumption for data communication. 
This paper presents CMSIS-NN, efficient kernels developed
to maximize the performance and minimize the memory footprint
of neural network (NN) applications on Arm Cortex-M processors targeted 
for intelligent IoT edge devices.
Neural network inference based on CMSIS-NN kernels achieves 4.6X improvement in 
runtime/throughput and 4.9X improvement in energy efficiency.
% for a 
%convolutional neural network targeting the CIFAR-10 dataset.
%Using these optimized neural network kernels, we demonstrate a
%convolutional neural network inference running on an off-the-shelf
%Cortex-M7 microcontroller with a throughput of 10.1 images per second
%(equivalently, 249 MOps per second) with 79.9\% accuracy.
\end{abstract}

\section{Introduction}
\label{sec:introduction}

Connected devices -- otherwise known as the Internet of Things (IoT) --
have been rapidly 
proliferating over the past few years and are predicted to reach 
1 trillion across various market segments by 2035~\cite{arm1tr}. 
These IoT edge devices typically consist of sensors collecting
data -- such as audio, video, temperature, humidity, GPS location
and acceleration -- which is then processed and communicated with 
other nodes or the cloud.
Currently, the data from the sensors are processed by 
analytics tools in the cloud to enable a wide range of applications, such as 
industrial monitoring and control, home automation  and
health care. However, as the number of the IoT nodes increases, this places a 
considerable burden on the network bandwidth, as well as adding latency to the
IoT applications.
Furthermore,  dependency on the cloud makes it challenging
to deploy IoT applications in regions with limited or unreliable
network connectivity. One solution to this problem is edge computing~\cite{edge}, 
performed right at the source of data, i.e. the IoT edge node, thus reducing 
latency as well as saving energy for data communication.

\begin{figure}[t]
\centering
\includegraphics[width = 0.8\columnwidth]{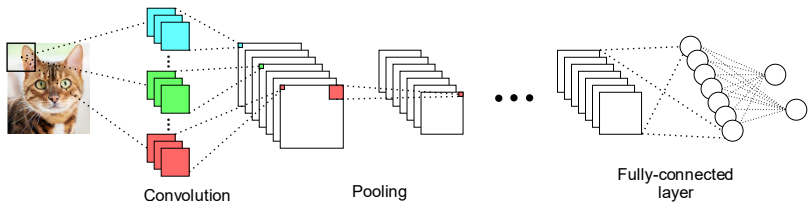}
\vspace{-11cm}
\caption{Structure of a typical deep neural network.}
\label{fig:cnn}
\end{figure}

In terms of accuracy, deep neural networks have demonstrated near-human
performance for many complex machine
learning applications such as image classification, speech 
recognition and natural language processing. 
A typical neural network (NN) for image classification consists of multiple layers 
of convolution based feature extractors, followed by fully-connected layers 
for classification, as shown in Fig.~\ref{fig:cnn}.
Due to the computational complexity and resource
requirements, the execution of NNs has predominantly been 
confined to cloud computing with high-performance server CPUs or specialized hardware 
(e.g. GPU or accelerators), which adds latency to the 
IoT applications. Classification using a small neural network 
right at the source of the data, i.e. the IoT edge, reduces the overall 
latency and energy consumption of data communication between the 
IoT edge and the cloud.

In this work, we explore the performance optimization of
neural networks on resource-constrained microcontroller based platforms, 
targeted for intelligent IoT edge nodes.
To enable this, we have developed optimized software
kernels for deploying NNs on Arm Cortex-M CPUs.
Using these software kernels, we demonstrate a convolutional neural
network (CNN) for CIFAR-10 dataset on an off-the-shelf Arm Cortex-M7 platform 
classifying 10.1 images per second with an accuracy of 79.9\%.

\section{Overview}
\label{sec:overview}

\begin{figure}[h]
\centering
\includegraphics[width = 0.65\columnwidth]{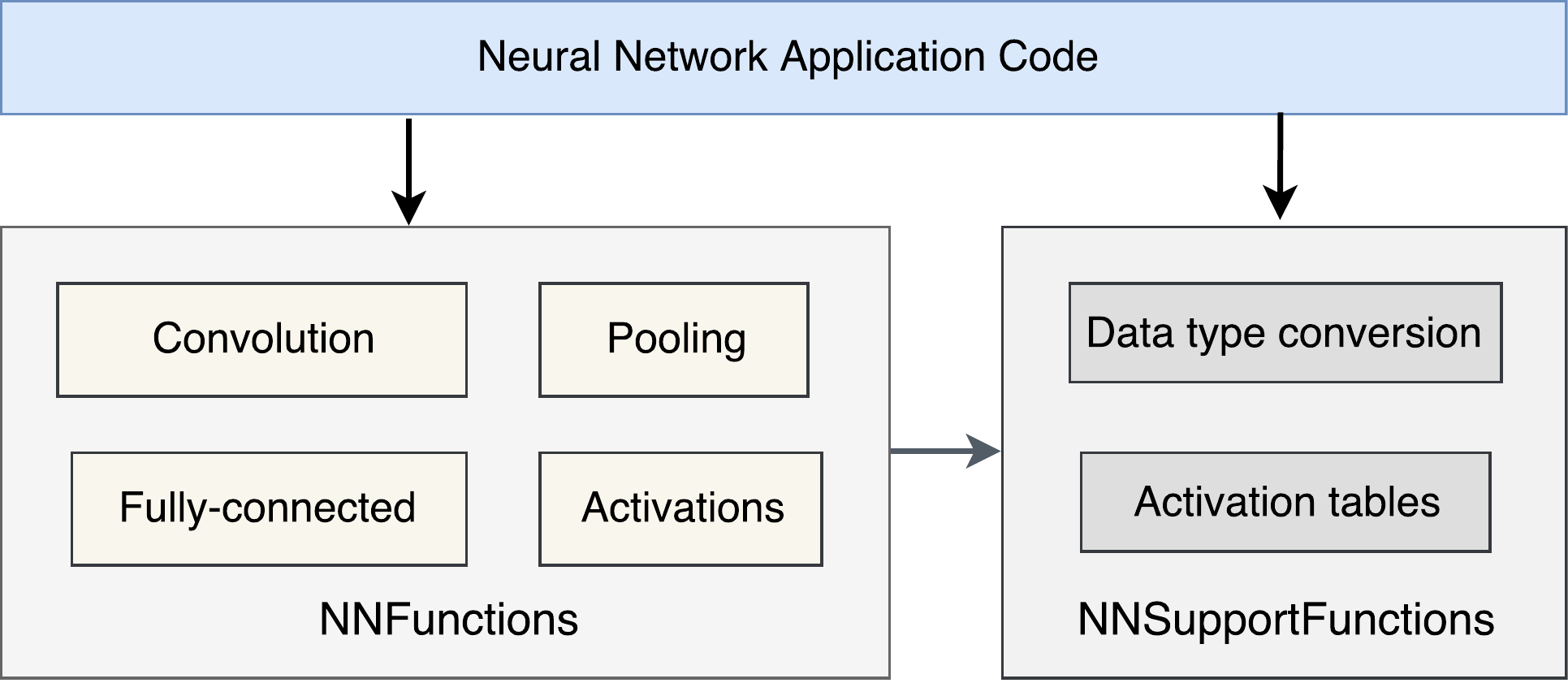}
\caption{Overview of the neural network kernel structure.}
\label{fig:overview}
\end{figure}

The overview of the neural network kernels is shown in Fig.~\ref{fig:overview}.
The kernel code consists of two parts: \textit{NNFunctions} and \textit{NNSupportFunctions}.
\textit{NNFunctions} include the functions that implement popular neural network layer 
types, such as convolution, depthwise separable convolution, fully-connected
(i.e. inner-product), pooling and activation. These functions can be used by the application 
code to implement the neural network inference applications. 
The kernel APIs are also kept simple, so that they can be easily retargeted for any machine 
learning framework.
\textit{NNSupportFunctions} include utility functions, such as
data conversion and activation function tables, which are used in
\textit{NNFunctions}. These utility functions can also be used by the application
code to construct more complex NN modules, such as Long Short Term Memory (LSTM) or Gated
Recurrent Unit (GRU).

For some kernels, such as fully-connected and convolution, different versions of
the kernel functions are implemented.
A basic version is provided that works universally, `as-is', for any layer 
parameters. We have also implemented other versions which include further 
optimization techniques with either transformed inputs or with some limitations 
on the layer parameters. Ideally, a simple script can be used to parse the
network topology and automatically determine the appropriate functions to be used.

\section{Fixed-Point Quantization}
\label{sec:quantization}

Traditionally, NN models are trained using 32-bit floating point data 
representation. However, such high precision is generally not required 
during inference.
Research has shown that NNs work well even with low-precision fixed-point 
representation~\cite{abadi2016tensorflow, lai2017deep, lin2016fixed}.
Fixed-point quantization helps to avoid the costly floating-point computation and
reduces the memory footprint for storing both weights and activations, 
which is critical for resource-constrained platforms.
Although precision requirements for different networks or network layers
can vary~\cite{suda2016throughput}, it is hard for the CPU to operate on data
types with varying bit-width. In this work, we develop the kernels that 
support both 8-bit and 16-bit data.

The kernels adopt the same data type format as used in CMSIS~\cite{cmsis}, 
i.e. $q7\_t$ as $int8$, $q15\_t$ as $int16$ and $q31\_t$ as $int32$.
The quantization is performed assuming a fixed-point format with a 
power-of-two scaling, i.e. the represented value will be
$A \times 2^{n}$, where $A$ is the integer value and
$n$ is an integer number that indicates the location of the radix point.
We pass the scaling factors for the bias and outputs as parameters to 
the kernels and the scaling is implemented as bitwise shift operations 
because of the power-of-two scaling.
We use this type of quantization -- instead of the 8-bit quantization
used in TensorFlow~\cite{abadi2016tensorflow} -- to 
avoid the need for floating-point de-quantization in between layers, as 
some Arm Cortex-M CPUs may not have a dedicated floating point unit (FPU),
thus limiting their floating-point computation capabilities.
The other benefit of such quantization is that we can use 
simpler table look-up based activation, which is discussed in Section~\ref{activation_sec}.

\section{Software Kernels}
\label{sec:kernels}

In this section, we describe the implementation and optimization of 
the proposed software kernels for Arm Cortex-M CPUs. 
The Cortex-M~\cite{cortex-m} family of processors are 32-bit RISC processor 
cores that are designed for energy efficiency, and typically used as microcontrollers
for deeply embedded applications.
In this work, we focus on enabling neural networks on Cortex-M based systems
that support SIMD instructions, especially 16-bit
Multiply-and-Accumulate (MAC) instructions (e.g. {\textit{SMLAD}}) which are very useful for
NN computation.

\subsection{Support Functions}
\label{subsec:q7_to_q15}
\begin{figure}[h]
\centering
\includegraphics[width = 0.95\columnwidth]{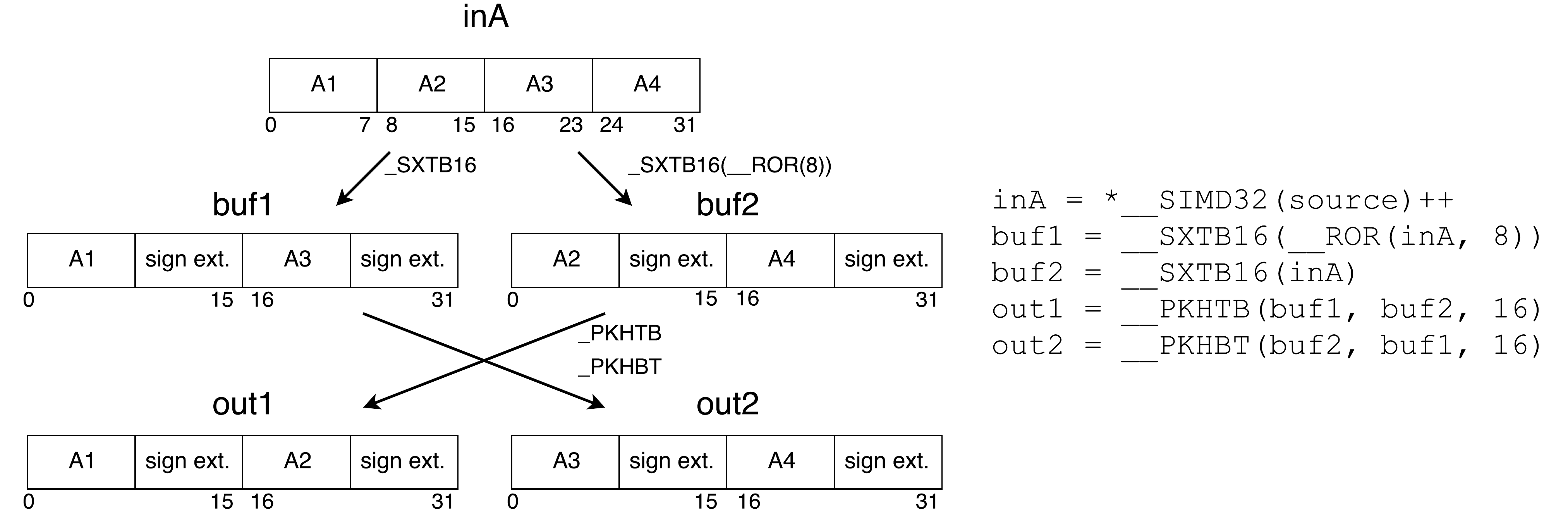}
\caption{Illustration and pseudo code of the data transform from $q7\_t$ to $q15\_t$ in CMSIS $arm\_q7\_to\_q15$ function
(assuming big-endian data format).}
\label{fig:q7_to_q15}
\end{figure}

Most \textit{NNFunctions} use the 16-bit MAC instructions, hence
data transformation is required to convert the 8-bit data type
(i.e. $q7\_t$) into 16-bit data type (i.e. $q15\_t$). 
CMSIS provides a utility function, $arm\_q7\_to\_q15$, to perform the
data transformation. The illustration and pseudo code is shown in Fig.~\ref{fig:q7_to_q15}.
The data transformation is done in two steps: the first step expands the
8-bit data into 16-bit data by using the sign extension instruction ($\_\_SXTB16$);
the second step rearranges the data so that the output follows
the same order as the input.

\begin{figure}[h]
\centering
\includegraphics[width = 0.95\columnwidth]{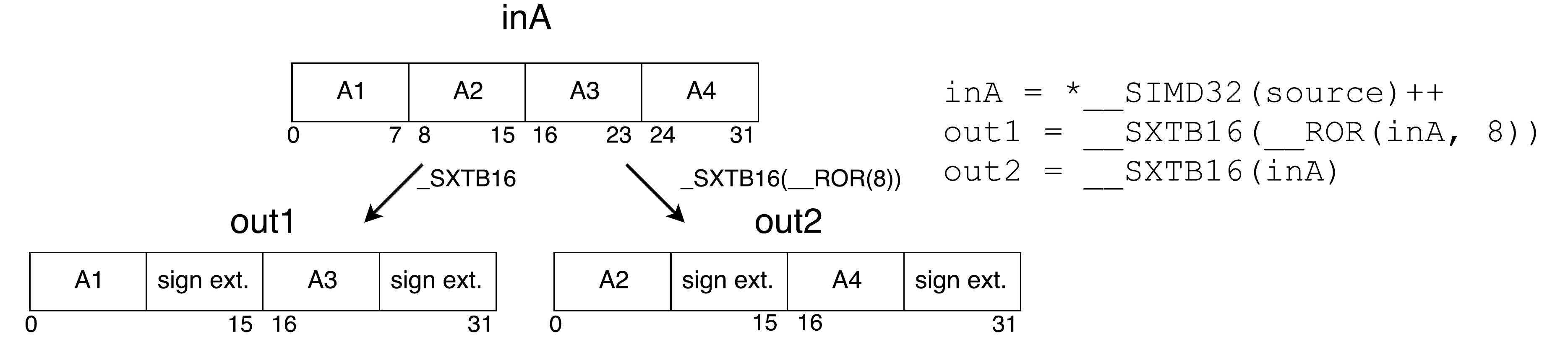}
\caption{Illustration and pseudo code for data transformation from $q7\_t$ to $q15\_t$ without reordering.
Output and input data are ordered differently.
}
\label{fig:q7_to_q15_no_shuffle}
\end{figure}

The performance of data transformation is critical, as it is used in the inner 
loop inside the computation kernels. While the first step of sign extension is
essential, the second step of rearranging the data can be omitted if both
operands follow the same ordering. To better exploit this, we created another
version of the data transformation routine without the data reordering,
as shown in Fig.~\ref{fig:q7_to_q15_no_shuffle}.
The routine is discussed in detail in Section~\ref{mat_mul_sec}.

\subsection{Matrix Multiplication}
\label{mat_mul_sec}
\begin{figure}[h]
\centering
\includegraphics[width = 0.75\columnwidth]{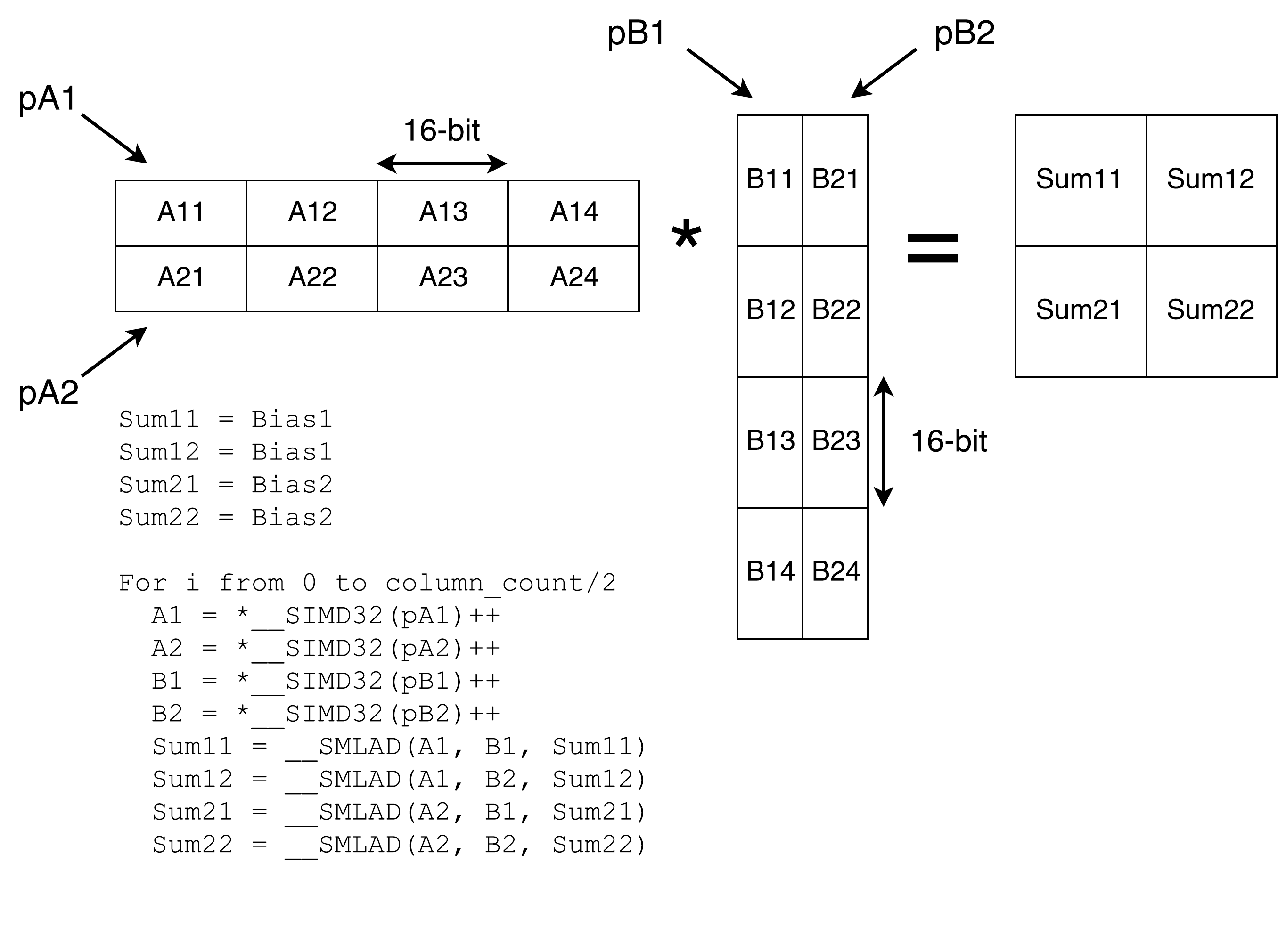}
\caption{The inner-loop of matrix multiplication with $2 \times 2$ kernel. Each
loop computes the dot product results of $2$ columns and $2$ rows, i.e.
$4$ outputs.}
\label{fig:mat_mult}
\end{figure}

Matrix multiplication is the most important computation kernel in
neural networks~\cite{gemmimportance}. The implementation in this work
is based on the $mat\_mult$ kernels in CMSIS.
Similar to CMSIS implementation, the matrix multiplication kernel 
is implemented with $2 \times 2$ kernels, illustrated in Fig.~\ref{fig:mat_mult}. 
This enables some data reuse and saves on the total number of load
instructions. The accumulation is done with the $q31\_t$ data type and
both operands are of $q15\_t$ data type. We initialize the accumulator
with the corresponding bias value. The computation is performed using
the dedicated MAC instruction $\_\_SMLAD$.

If the input activations or network weights are $q7\_t$ type,
data expansion may be required to convert them to $q15\_t$ type.
As discussed in Section ~\ref{subsec:q7_to_q15},
if both inputs and weights are $q7\_t$ type, we can use data
transformation without reordering (as shown in Fig.~\ref{fig:q7_to_q15_no_shuffle})
to improve performance.
However, the data alignment can be tricky when the number of elements
is not a multiple of $4$.

The other scenario is with $q7\_t$ weights and $q15\_t$ activations.
In this case, the weights can be pre-processed with the second
and third byte swapped for every 32-bit word, i.e. converting $[1,2,3,4]$
into $[1,3,2,4]$. 
With this pre-processing,
the data transformation without reordering (as shown in Fig.~\ref{fig:q7_to_q15_no_shuffle})
will generate $q15\_t$ data in the original order, i.e. converting $[1,3,2,4]$
back to $[1,2,3,4]$.
This pre-processing is only for the network weights and can be
reused for different inputs.
Alternatively, the pre-processing can be performed offline when generating the
network model.

We consider the fully-connected layer to have a batch size of one, so the main
computation becomes matrix-vector multiplication. Similar to 
matrix-matrix multiplication, matrix-vector multiplication 
can also be implemented with a $1 \times 2$ kernel size to improve
performance. Supporting a large kernel can further improve
performance, but may be limited by the total number of
registers. Arm Cortex-M cores have $16$ architectural registers,
including {\textit{PC}} and {\textit{LR}}. This limited number of registers 
can be a challenge if we want to implement larger kernels.

\begin{figure}
\centering
\includegraphics[width = 0.70\columnwidth]{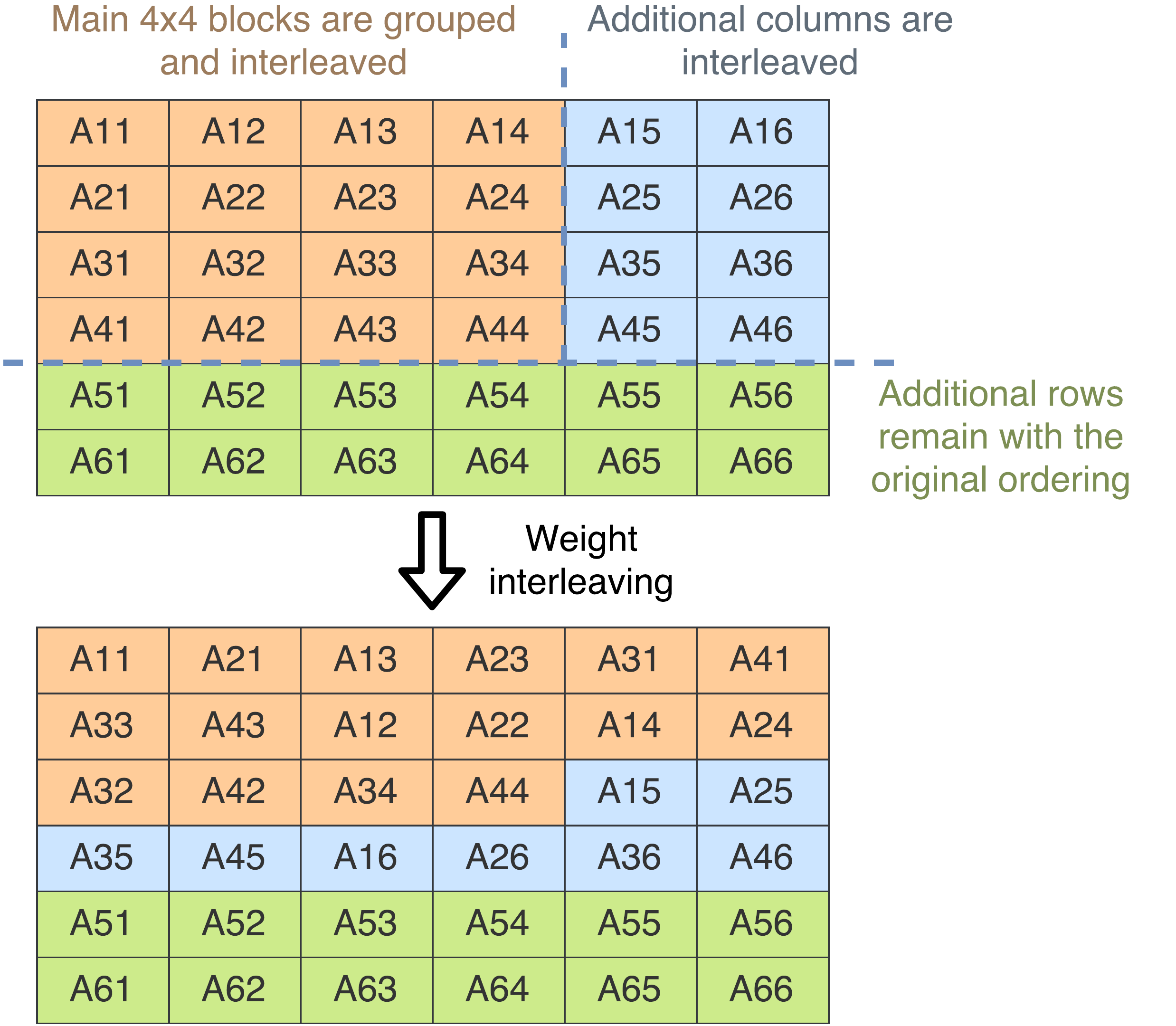}
\caption{The weight reordering process to support a $1 \times 4$ kernel 
(i.e., $1$ column and $4$ rows).
Weights are interleaved every four rows and shuffled every four
entries in the main part. Only the leftover columns are interleaved, whereas
leftover rows remain the same.}
\label{fig:x4_reordering}
\end{figure}

\begin{figure}
\centering
\includegraphics[width = 0.95\columnwidth]{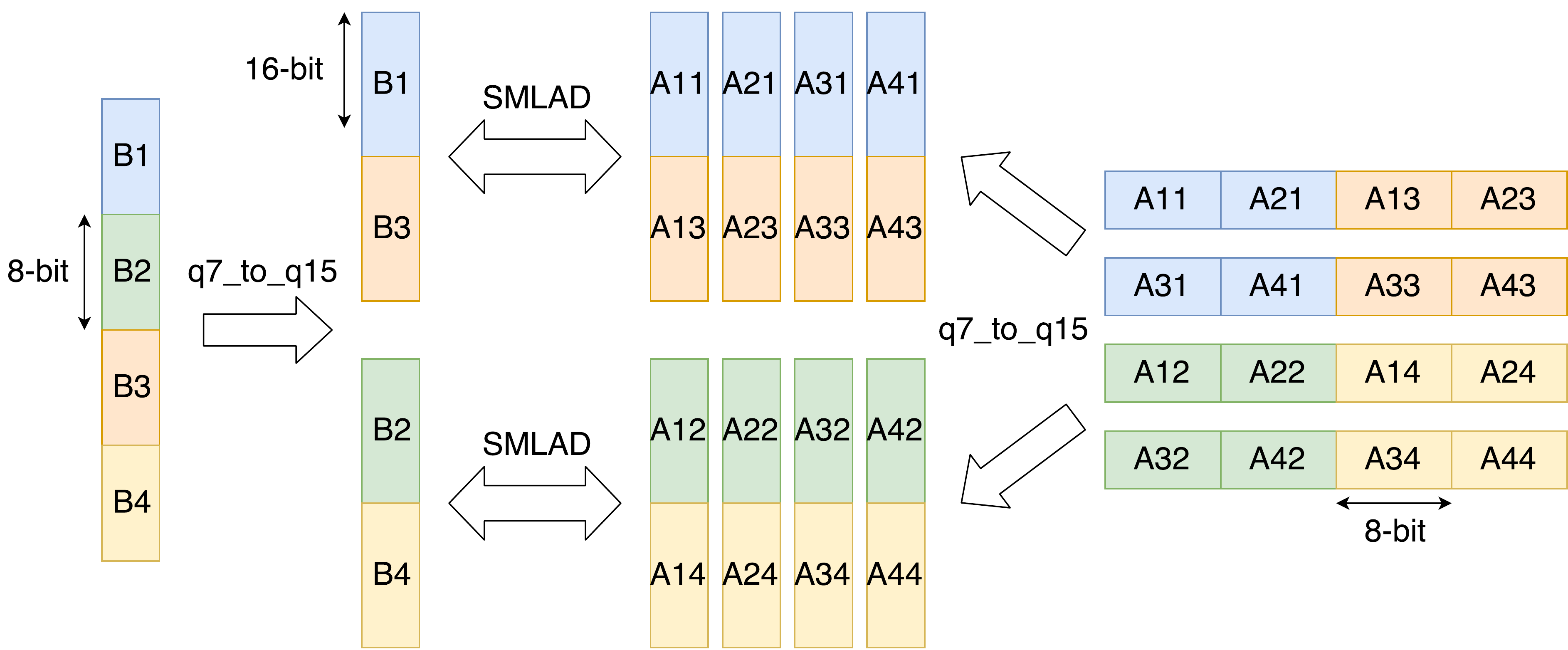}
\caption{The computation of a $1 \times 4$ kernel. The data order of the weights
and activation vectors is matched after the $q7\_to\_q15$ operation without
reordering. Each inner loop iteration processes two $1 \times 4$ MAC operations.}
\label{fig:x4_computation}
\end{figure}

Since the weights are kept constant and re-used during inference,
we can reorder the matrix weights so that row data are interleaved
and can be read with only one pointer access.
This weight reordering is illustrated in Fig.~\ref{fig:x4_reordering}.
During the matrix-vector multiplication, the same $q7\_to\_q15$ function, 
without reordering, is used to expand the $q7\_t$ data into
$q15\_t$ as shown in Fig.~\ref{fig:x4_computation}. In this way, we can
fit the $1 \times 4$ kernels using the available registers.

\subsection{Convolution}
\begin{figure}[h]
\centering
\includegraphics[width = 0.85\columnwidth]{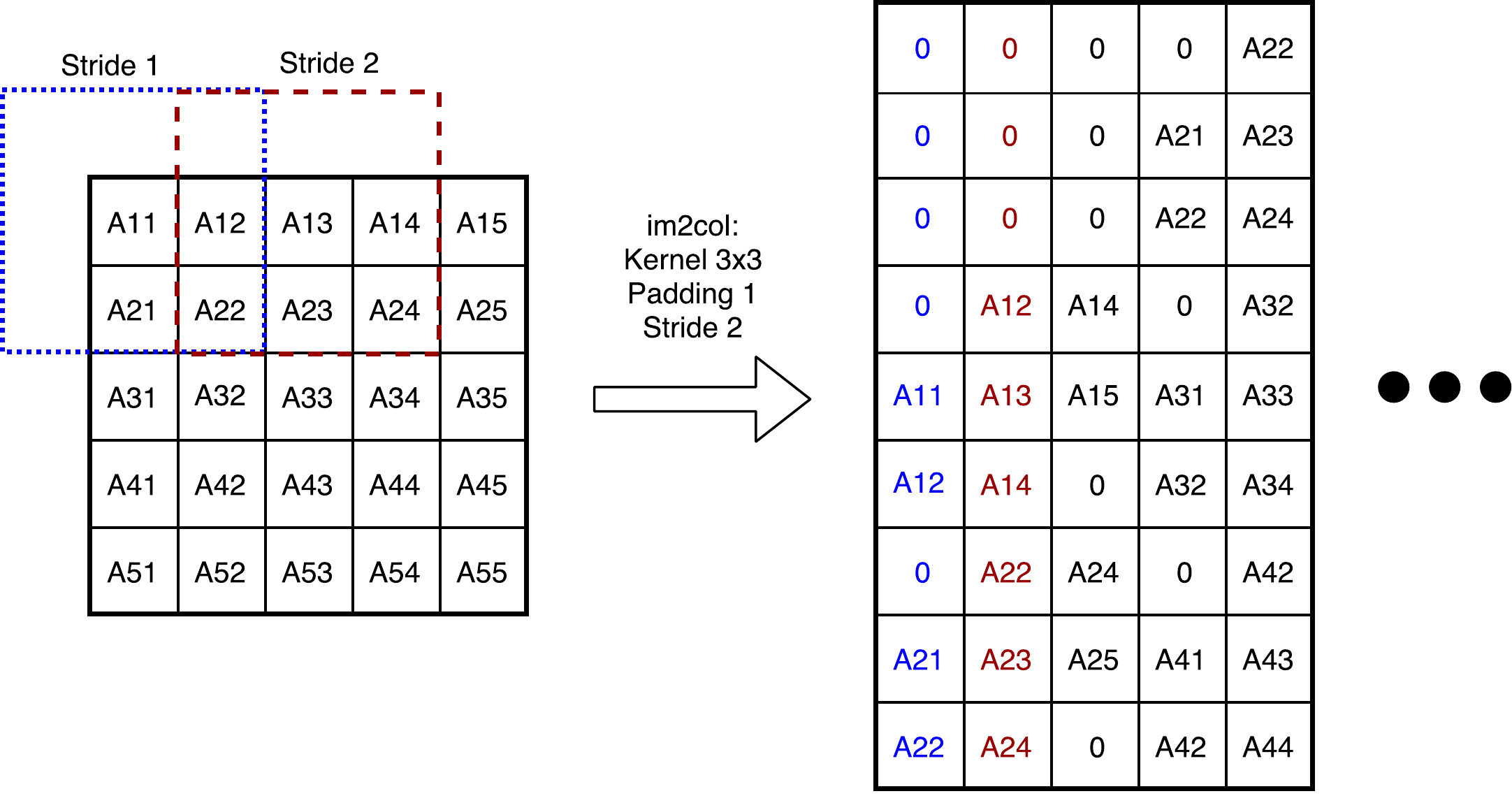}
\caption{Example of \textit{im2col} on a 2D image with a 3x3 kernel, 
padding size of 1 and stride size of 2.}
\label{fig:im2col}
\end{figure}

A convolution layer extracts a new feature map by computing a dot product 
between filter weights and a small receptive field in the input 
feature map.
Typically, a CPU-based implementation of convolution is decomposed into
input reordering and expanding (i.e. {\textit{im2col}}, image-to-column)
and matrix multiplication operations. 
\textit{im2col} is a process of transforming the image-like
input into columns that represent the data required by each convolution filter.
An example of \textit{im2col} is shown in Fig.\ref{fig:im2col}.

One of the main challenges with \textit{im2col} is the increased memory footprint, since the pixels 
in the input image are repeated in the \textit{im2col} output matrix.
To alleviate the memory footprint issue while retaining the performance benefits
from \textit{im2col}, we implemented a \textit{partial im2col} for our convolution kernels.
The kernel will only expand a limited number of columns (e.g. 2), sufficient
to get the maximum performance boost from the matrix-multiplication kernels while
keeping memory overhead minimal.

\begin{figure}[h]
\centering
\includegraphics[width = 0.65\columnwidth]{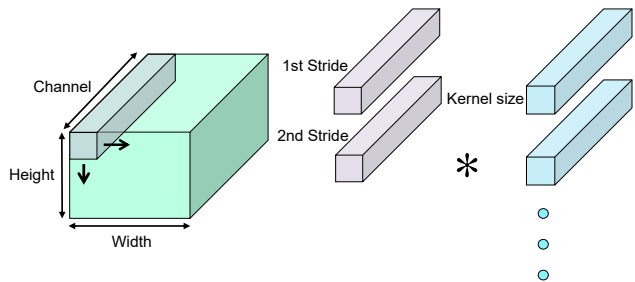}
\vspace{-8cm}
%\caption{Illustration of convolution on 3D data. The image has three dimensions: height, width and channel.}
\caption{Convolution on 3D data. The image has three dimensions: height, width and channel.}
\label{fig:hwc_example}
\end{figure}

The image data format can also affect the performance of convolution,
especially \textit{im2col} efficiency~\cite{li2016optimizing}.
With a batch size of one, the convolution operation is a 2D convolution 
(i.e. the convolution window can move in two directions) on 3D data,
as shown in Fig.~\ref{fig:hwc_example}. 
The two most common image data formats are Channel-Width-Height (CHW), i.e. channel last,
and Height-Width-Channel (HWC), i.e. channel first.
The dimension ordering is the same as that of the data stride. 
In an HWC format, the data along the channel is stored with a stride of 1,
data along the width is stored with a stride of the channel count,
and data along the height is stored with a stride of (channel count $\times$ image width).

\begin{figure}[h]
\centering
\includegraphics[width = 0.65\columnwidth]{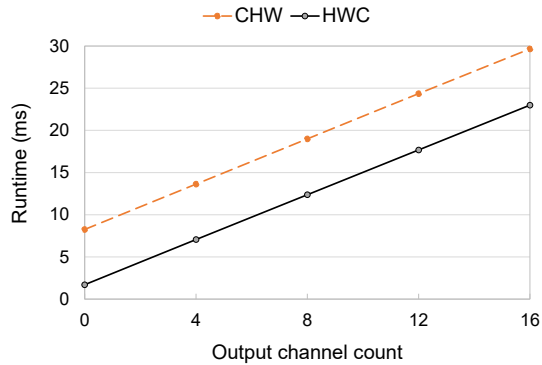}
\vspace{-6.5cm}
\caption{Experiment results with CHW and HWC data layout. Both data layout styles have the same
matrix-multiplication runtime. HWC has less \textit{im2col} runtime.}
\label{fig:hwc_vs_chw}
\end{figure}

The data layout has no impact on the matrix-multiplication operations, as long as 
the dimension order of both weights and images is the same.
The \textit{im2col} operations are performed along the width 
and height dimensions only. The HWC-style layout enables efficient data movement, 
as data for each pixel (i.e. at the same x,y location) is stored contiguously
and can be copied efficiently with SIMD instructions. 
To validate this, we implemented both CHW and HWC versions and compared the runtime
on a Cortex-M7. The results are highlighted in Fig.~\ref{fig:hwc_vs_chw}, where we fixed the HWC input
to be 16x16x16 and swept the number of output channels. When the output channel value is zero,
it means that the software performs only \textit{im2col} and no matrix-multiplication operation.
Compared to CHW layout, HWC has less \textit{im2col} runtime with the same matrix-multiplication
performance. Therefore, we implement the convolution kernels assuming that the data layout
is in HWC format.

Recently, MobileNets using depthwise separable convolution have been proposed as an efficient 
alternative to the standard 3D convolution operation~\cite{howard2017mobilenets}. 
Depthwise separable convolution has been used to achieve 
compact network architectures, which are particularly useful for resource-constrained 
devices~\cite{zhang2017hello}. CMSIS-NN kernels also include support for a depthwise separable
convolution layer.

\subsection{Pooling}

Pooling layers are typically inserted in between convolution layers
to reduce the feature dimensions and thus the number of parameters
and computations in the network. Similar to convolution, pooling is a window-based 
operation with a given kernel size, stride and padding. Unlike convolution, pooling typically
operates within the same channel, and is independent of data in the other channels.
Pooling is usually performed with a stride size greater than 1, so the output features
will have smaller width and height.

There are two common types of pooling layers: average pooling, which calculates
the average value of all pixels within the window, and max pooling, which calculates
the maximum value. 
Pooling can be implemented as a nested for-loop over each window,
e.g. pooling layers in Caffe~\cite{jia2014caffe}. 
One efficient alternative is to split the pooling operation into x-pooling
(i.e. along the width) and then y-pooling (i.e. along the height). 
This way, the max/average operations along the x-direction can be reused along the
y-direction, allowing the total number of operations to be reduced. 
We call this approach split x-y pooling. One potential issue with split x-y pooling
is the data arrangement, as additional memory may be required to store 
the intermediate results after x-pooling.

\begin{figure}[h]
\centering
\includegraphics[width = 0.75\columnwidth]{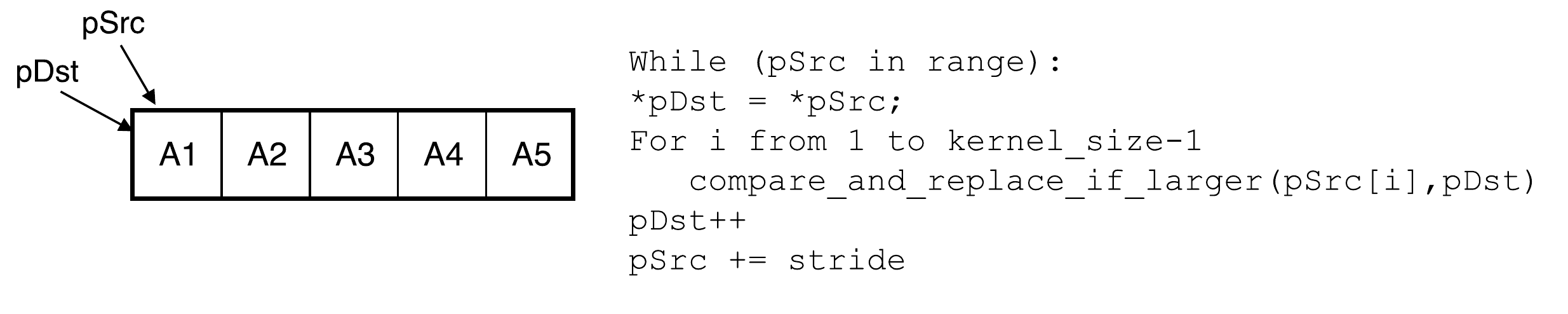}
\caption{Example of max pooling with \textit{in situ} operations.}
\label{fig:pooling}
\end{figure}

Our pooling kernels are implemented with split x-y pooling as, based on our experiments, 
the split x-y pooling is significantly faster than the window-based pooling.
To eliminate the need for additional memory, the kernels perform the pooling
operations \textit{in situ}. This makes the pooling layer a destructive operation on 
the input.
An example of an \textit{in situ} max pooling implementation on a 1D array 
is illustrated in Fig.~\ref{fig:pooling}.
The same operation can be extended for high dimensional data where each element
can represent an array. For example, when performing x-pooling with HWC image data,
each element block in Fig.~\ref{fig:pooling} can represent an array of size
equal to the number of channels, i.e. all the channel data for one pixel.
If each block represents an entire image row, this will effectively perform 
y-pooling.
Compared to window-based pooling, the \textit{in situ} split x-y pooling achieves 4.5X speed-up 
with no additional memory overhead.

\subsection{Activation Functions}
\label{activation_sec}
The role of an activation function is to add non-linearity in the network. The most commonly
used activations functions are ReLU, sigmoid and tanh.

\subsubsection{ReLU}

\begin{figure}[h]
\centering
\includegraphics[width = 0.85\columnwidth]{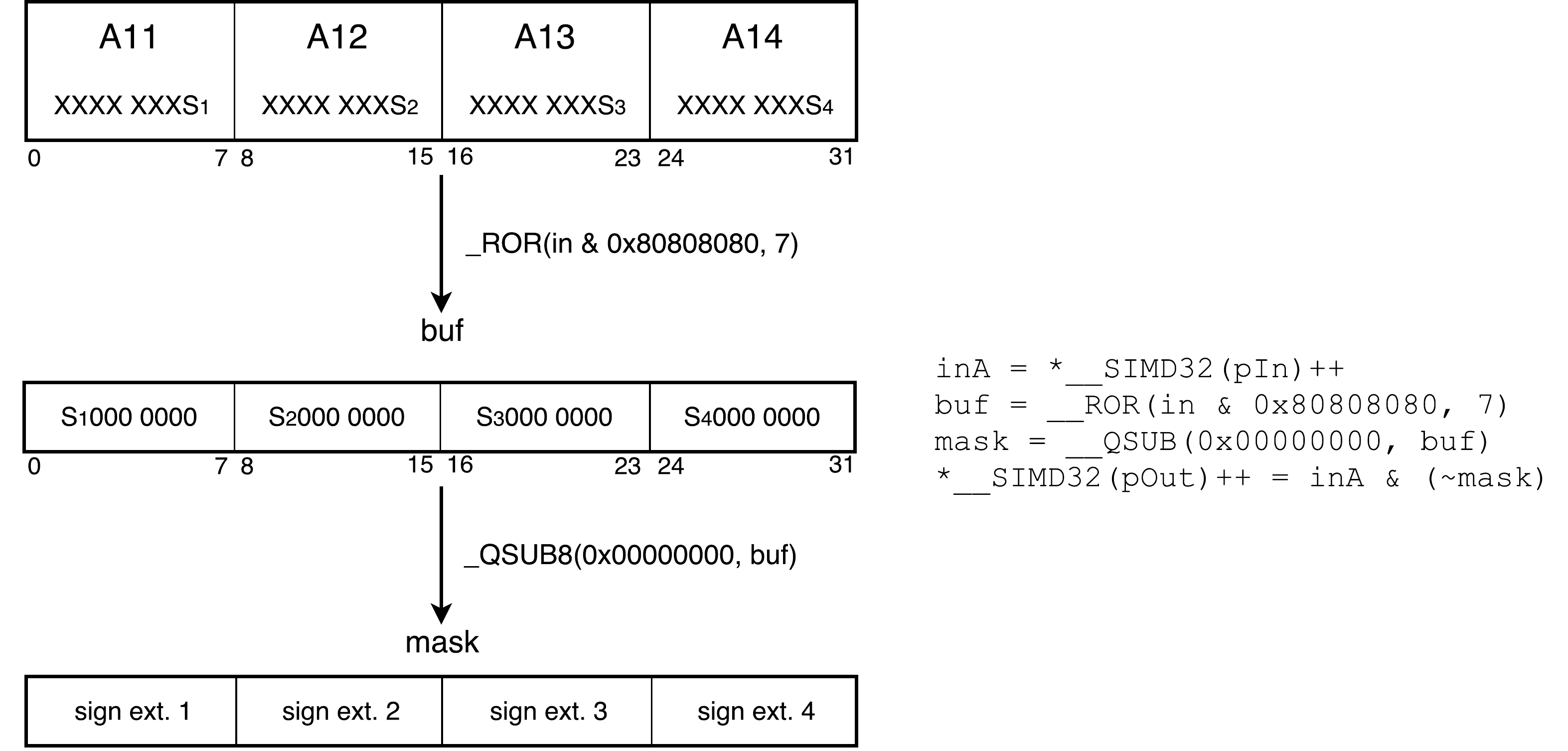}
\caption{Illustration of the optimized ReLU kernel. The MSB of each $q7\_t$ data
is used to construct a data mask. If the number is negative, i.e. the sign bit is 1,
the mask is 0xFF, which will make the output 0.}
\label{fig:relu}
\end{figure}

Rectified-Linear Unit (ReLU) layer is a common activation function for neural networks. The operation is
$f(x) = max(0, x)$, which is $0$ when $x<0$ and linear with a slope of $1$ when $x>0$.
A simple implementation (e.g. in Caffe) loops over all elements and make
them 0 if they are negative.
Since we mostly use $q7\_t$ type data, ReLU can be implemented using a similar concept as
SWAR (SIMD within a register). The key is to identify the sign bit of the $q7\_t$ number and 
make the number 0 if it is negative. 
Our ReLU kernel implementation is shown in Fig.~\ref{fig:relu}. The basic idea is to use
the MSB of the $q7\_t$ number as the sign bit and extend it into a mask by using the byte-level
subtraction instruction ($\_\_QSUB8$). This SWAR approach provides about 4X speed-up compared to 
the conventional approach of going over each element.

\subsubsection{Sigmoid and Tanh}

Sigmoid and tanh are other common activation functions that are typically used in 
recurrent neural networks. 
Computing these activation functions requires the use of dedicated math functions and 
can be computationally expensive on Cortex-M CPUs, hence we implement them using 
a table-lookup approach  with fixed-point input and output. 
There are two possible ways to do this, the first of which is to use a unified table for all ranges 
with fixed input granularity. In this implementation,
the MSB of the input is used to identify the corresponding entries in the look-up table,
while the LSB can be used for linear interpolation, if needed. This is similar to the sine 
and cosine table look-up in CMSIS.

The other option is to implement two separate tables to cover different regions of the
functions. This can improve accuracy, as both sigmoid and tanh functions are highly
non-linear. Ideally, there should be a table with finer granularity for an input region around
$0$ and another table with coarser granularity for inputs with larger absolute values.
In this case, the function first determines which table to use based on the input
data range, e.g. by looking at the MSB. After that, similar to the unified table approach,
the LSB can be used for calculating the table entries and interpolation.

Unlike periodic functions such as sine or cosine, it is important to determine the
table range for sigmoid and tanh functions. We determine that range of $[-8, 8]$
works well enough for both functions, as $sigmoid(8) = 0.9997$ and $tanh(8) = 0.9999$.
%For some networks, the activation may also have limited range. 
We can also
generate other versions of the look-up table with different ranges, e.g. $[-4, 4]$.

\section{Experimental Results}

We tested the CMSIS-NN kernels on a CNN trained on the CIFAR-10 dataset, consisting of 
60,000 32x32 color images divided into 10 output classes. 
The network topology is based on the built-in example provided in Caffe, with three convolution
layers and one fully-connected layer. All the layer weights and activation data are
quantized to $q7\_t$ format. The layer parameters and the detailed runtime results using 
the CMSIS-NN kernels are shown in the Table~\ref{tbl:network}. 
The runtime is measured on a NUCLEO-F746ZG Mbed board~\cite{nucleo_m7} with an
Arm Cortex-M7 core running at 216 MHz. 

\begin{table}[h]
\setlength\extrarowheight{2pt}
\caption{Layer parameters and performance for the CIFAR-10 CNN.}
\label{tbl:network}
\begin{center}
\begin{tabular}{|c|c|c|c|c|c|}
\hline
        & Layer Type  & Filter Shape       & Output Shape       & Ops &  Runtime   \\
\hline 
Layer 1 & Convolution & 5x5x3x32 (2.3 KB) &  32x32x32 (32 KB)  & 4.9 M                &  31.4 ms    \\
\hline
Layer 2 & Max Pooling &  N.A.              &  16x16x32 (8 KB)   & 73.7 K               &  1.6 ms  \\
\hline
Layer 3 & Convolution & 5x5x32x32 (25 KB)  &  16x16x32 (8 KB)   & 13.1 M               &  42.8 ms \\
\hline
Layer 4 & Max Pooling &  N.A.              &  8x8x32 (2 KB)     & 18.4 K               &  0.4 ms  \\
\hline
Layer 5 & Convolution & 5x5x32x64 (50 KB)  &  8x8x64 (4 KB)     & 6.6 M                &  22.6 ms  \\
\hline 
Layer 6 & Max Pooling &  N.A.              &  4x4x64 (1 KB)     & 9.2 K                &  0.2 ms \\
\hline
Layer 7 & Fully-connected & 4x4x64x10 (10 KB) & 10              & 20 K                 &  0.1 ms \\
\hline
Total   &                 & 87 KB weights  &  55 KB activations & 24.7 M               &  99.1 ms \\
\hline
\end{tabular}
\end{center}
\end{table}

The entire image classification takes about 99.1 ms per image (the equivalent of
10.1 images per second). The compute throughput of the CPU is about 249 MOps per second 
for running this network. The pre-quantized network achieves an accuracy of 
80.3\% on the CIFAR-10 test set. The 8-bit quantized network running on Arm Cortex-M7 core
achieves 79.9\% accuracy. Maximum memory footprint using the 
CMSIS-NN kernels is $\sim$133 KB, where convolutions are implemented with \textit{partial im2col} 
to save memory, followed by matrix-multiplication. Memory footprint without partial 
\textit{im2col} would be $\sim$332 KB and the neural network would not fit on the board.

\begin{table}[b]
\setlength\extrarowheight{2pt}
\caption{Throughput and energy efficiency improvments by layer types}
\label{tbl:layers}
\begin{center}
\begin{tabular}{|c|c|c|c|c|}
\hline
\multirow{2}{*}{Layer type}  &  \multirow{2}{*}{Baseline runtime}  &  \multirow{2}{*}{New kernel runtime}  &
\multicolumn{2}{c|}{Improvement} \\
\cline{4-5}
              &            &              & Throughput  & Energy Efficiency \\
\hline
Convolution   &  443.4 ms  &  96.4 ms     &    4.6X     &   4.9X   \\
\hline
Pooling       &  11.83 ms  &  2.2 ms      &    5.4X     &   5.2X   \\
\hline
ReLU          &   1.06 ms  &  0.4 ms      &    2.6X     &   2.6X   \\
\hline
Total         &   456.4ms  &  99.1 ms     &    4.6X     &   4.9X   \\
\hline
\end{tabular}
\end{center}
\end{table}

To quantify the benefits of CMSIS-NN kernels over existing solutions, we also implemented a 
baseline version using
a 1D convolution function ($arm\_conv$ from CMSIS-DSP), Caffe-like pooling and ReLU.
For the CNN application, Table~\ref{tbl:layers} summarizes the comparison results of the baseline
functions and the CMSIS-NN kernels. The CMSIS-NN kernels achieve 
2.6X to 5.4X improvement in runtime/throughput over the baseline functions. The energy efficiency 
improvement is also in line with the throughput improvement.

\section{Conclusion}

We developed CMSIS-NN to maximize the performance and minimize the memory footprint of 
neural networks on Arm Cortex-M CPUs.
Neural network inference based on CMSIS-NN kernels achieved 4.6X improvement in 
runtime/throughput and 4.9X improvement in energy efficiency for a 
convolutional neural network targeting the CIFAR-10 dataset.
%Using these optimized neural network kernels, we demonstrated a
%CNN inference running on an off-the-shelf
%Cortex-M7 microcontroller with a throughput of 10.1 images per second
%(equivalently, 249 MOps per second) with 79.9\% accuracy.
The CMSIS-NN kernels are available at \url{https://github.com/ARM-software/CMSIS\_5}.
The application code can directly use these kernels to implement neural network 
algorithms on Arm Cortex-M CPUs. Alternatively, these kernels can be used 
as primitives by machine learning frameworks to deploy trained models.

%\bibliography{paper}
\bibliographystyle{unsrt}

\end{document}